\pdfoutput=1

\documentclass[11pt]{article}

\usepackage{CJKutf8}
\usepackage[preprint]{acl}

\usepackage{times}
\usepackage{latexsym}

\usepackage[T1]{fontenc}

\usepackage[utf8]{inputenc}

\usepackage{microtype}
\usepackage{textcomp}

\usepackage{inconsolata}
\usepackage{graphicx}
\usepackage{subfig}

\usepackage{amsmath}
\usepackage{booktabs}

\usepackage{tikz,pgfplots}

\definecolor{nyuviolet}{RGB}{87,6,140}
\definecolor{nyulightviolet1}{RGB}{171,130,197}
\definecolor{nyulightviolet2}{RGB}{238,230,243}
\definecolor{nyuteal}{RGB}{0,156,139}

\title{ERAS: Evaluating the Robustness of Chinese NLP Models to Morphological Garden Path Errors}

\author{Qinchan (Wing) Li \\
  Tandon School of Engineering\\
  New York University \\
  \texttt{ql840@nyu.edu} \\\And
  Sophie Hao \\
  Center for Data Science \\
  New York University \\
  \texttt{sophie.hao@nyu.edu}}

\pgfplotsset{compat=1.18}

\begin{document}
\begin{CJK}{UTF8}{gbsn}
\maketitle
\begin{abstract}
    In languages without orthographic word boundaries, NLP models perform \textit{word segmentation}, either as an explicit preprocessing step or as an implicit step in an end-to-end computation. 
    This paper shows that Chinese NLP models are vulnerable to \textit{morphological garden path errors}---errors caused by a failure to resolve local word segmentation ambiguities using sentence-level morphosyntactic context. We propose a benchmark, \textit{ERAS}, that tests a model's vulnerability to morphological garden path errors by comparing its behavior on sentences with and without local segmentation ambiguities. Using ERAS, we show that word segmentation models make garden path errors on locally ambiguous sentences, but do not make equivalent errors on unambiguous sentences. We further show that sentiment analysis models with character-level tokenization make \textit{implicit} garden path errors, even without an explicit word segmentation step in the pipeline. Our results indicate that models' segmentation of Chinese text often fails to account for morphosyntactic context.
\end{abstract}

\section{Introduction}

When working with languages that do not mark word boundaries in their writing systems, the task of \textit{word segmentation}, where texts are parsed into word-level tokens, is non-trivial. Nonetheless, in high-resource languages like Chinese, the state of the art in word segmentation is strong, with F1 scores close to 100\% for common benchmarks \citep[\textit{inter alia}]{emersonSecondInternationalChinese2005,huangFastAccurateNeural2020,keUnifiedMultiCriteriaChinese2020,linImprovingMultiCriteriaChinese2023}. %

Word segmentation is subject to structural ambiguity: there may be more than one way to segment a text into words or morphemes. In English, for example, the word \textit{unlockable} can be understood to mean `unable to be locked' (\textit{un} + \textit{lockable}), or `able to be unlocked' (\textit{unlock} + \textit{able}). In some cases, a sequence of words or morphemes that appears ambiguous may be disambiguated by morphosyntactic context. \autoref{fig:gardenpath-example} illustrates this \textit{morphological garden path} phenomenon with an example from Mandarin Chinese. In isolation, the trigram 留心机 could be segmented as either 留 `with' + 心机 `caution' or 留心 `take note of' + 机 `the machine'. In context, however, the rest of the sentence cannot be parsed grammatically if the latter segmentation is used. We therefore say that this sentence is \textit{locally ambiguous}---the sentence as a whole is structurally unambiguous, but it contains a substring that is ambiguous in isolation.

\begin{figure}
    \centering
    \footnotesize
    \textbf{Grammatically Valid Segmentation}
    
    \vspace{1em}
    \begin{tabular}{c c c c c}
        学生 & \underline{留} & \underline{心机} & 处理 & 友人  \\
        the student & with & caution & treats & his friends
    \end{tabular} 
    \vspace{.25em}
    
    `The student treats his friends with caution.'

    \vspace{1em}
    \textbf{Ungrammatical Segmentation}
    
    \vspace{1em}
    \begin{tabular}{c c c c c}
        学生 &  \colorbox{nyuviolet!30}{\underline{留心}} &\underline{机} & 处理 & 友人 \\
        the student & takes & the & dealing & his friends \\
        & note of & machine & with
    \end{tabular}
    
    \caption{Morphological garden paths involve local ambiguity at the level of word segmentation. In the sentence above, the bigram \colorbox{nyuviolet!30}{留心} constitutes a valid word, but segmenting it as a word renders the sentence unparsable. A valid parse is obtained if the character 心 forms a word with the following character, 机.}
    \label{fig:gardenpath-example}
\end{figure} 

\begin{figure}
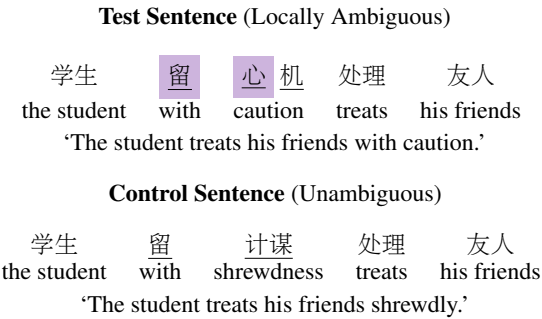

    \centering
    \footnotesize
    \textbf{Test Sentence} (Locally Ambiguous)
    
    \vspace{1em}
    \begin{tabular}{c c c c c }
        学生 & \colorbox{nyuviolet!30}{\underline{留}} &\colorbox{nyuviolet!30}{\underline{心}} \underline{机} & 处理 & 友人  \\
        the student & with & caution & treats & his friends
    \end{tabular} 
    \vspace{.25em}
    
    `The student treats his friends with caution.'

    \vspace{1em}
    \textbf{Control Sentence} (Unambiguous)
    
    \vspace{1em}
    \begin{tabular}{c c c c c }
        学生 & \underline{留} & \underline{计谋} & 处理 & 友人  \\
        the student & with & shrewdness & treats & his friends \\
    \end{tabular} 
    \vspace{.25em}
    
    `The student treats his friends shrewdly.'
    
    \caption{ERAS consists of locally ambiguous \textit{test sentences} paired with unambiguous \textit{control sentences}. Test and control sentences differ in terms of a three-character \underline{test site}, where test sentences contain a two-character \colorbox{nyuviolet!30}{canary word} whose existence renders the sentence unparsable.}
    \label{fig:eras}
\end{figure}

This paper investigates whether Chinese NLP models correctly resolve local structural ambiguities caused by morphological garden paths. If they do not, then we expect models to make \textit{garden path errors} by parsing locally ambiguous substrings incorrectly. To that end, we propose the \underline{E}valuation of \underline{R}obustness to Locally \underline{A}mbiguous \underline{S}egmentation (ERAS) benchmark, a synthetic test set consisting of locally ambiguous \textit{test sentences} paired with unambiguous but otherwise identical \textit{control sentences}. As illustrated in \autoref{fig:eras}, each test sentence contains a \textit{canary word} that can only appear in a segmentation of the sentence if a locally ambiguous substring is parsed incorrectly.

ERAS can be used in two ways. First, ERAS can evaluate word segmentation models by comparing the models' segmentations for test and control sentences. If a canary word is parsed as a word, but the equivalent segmentation error is not made for the control sentence, then we assume that a garden path error has been committed. ERAS can also evaluate sentiment analysis models that use a byte- or character-level tokenization scheme. Using a lexicon containing human-annotated word-level sentiment ratings \citep{wangANTUSDLargeChinese2016}, we ensure that canary words carry a sentiment polarity that is incongruent with the rest of the sentence. If the model's predicted sentiment for the test sentence is influenced by the sentiment of the canary word, then we assume that an \textit{implicit} garden path error has been committed.

This paper conducts both types of evaluations using ERAS. We first show that Transformer-based \citep{vaswaniAttentionAllYou2017,devlinBERTPretrainingDeep2019} and non-neural word segmentation models make garden path errors, systemically parsing canary words as words (\autoref{sec:word-segmentation}). We show that these errors are mostly caused by a preference for resolving garden paths in a greedy left-to-right manner. We then show that Transformer-based sentiment analysis models with a character-level tokenization scheme make implicit garden path errors on ERAS (\autoref{sec:sentiment}). These errors can be mitigated by introducing word boundary information to the model during pre-training or fine-tuning. Our findings point to morphological garden paths as a source of error for Chinese NLP models, especially for models that do not receive any explicit word boundary information during pre-training or fine-tuning.

\paragraph{Contributions.} (1) We provide a benchmark, ERAS, for determining whether a Chinese NLP model explicitly or implicitly resolves local ambiguities due to morphological garden paths. (2) Using ERAS, we show that word segmentation models are vulnerable to explicit garden path errors, and sentiment analysis models are vulnerable to implicit garden path errors.

\section{ERAS: A Benchmark for Detecting Garden Path Errors}

ERAS consists of 203,944 pairs of test and control sentences, available in both simplified and traditional characters, synthetically generated from templates. ERAS is organized into 39 \textit{minimal pair paradigms} \citep{warstadtBLiMPBenchmarkLinguistic2020}, examples of which are shown in \autoref{fig:paradigms}. Each paradigm is defined by two templates, one for test sentences and one for control sentences. Within each paradigm, the two templates are identical, except that the test template contains a morphological garden path and the control template does not. 

We evaluate models on ERAS by testing the null hypothesis that models perform similarly on test and control sentences. If a model performs better on control sentences than on test sentences, then we conclude that this model is susceptible to garden path errors. ERAS is designed to test two kinds of models: Chinese word segmentation models, which may incorrectly segment morphological garden paths, and sentiment analysis models, whose outputs may be influenced by the sentiment of words that could only exist if morphological garden paths are implicitly segmented incorrectly.

\subsection{Dataset Structure}

As shown in Figures \ref{fig:eras} and \ref{fig:paradigms}, test and control templates differ in terms of a three-character substring called the \textit{test site}. In test templates, the first two and last two characters of the test site form words from the ANTUSD sentiment lexicon, which overlap at the middle character. One of the two words is designated as the \textit{true word}, while the other is designated as the \textit{canary word}. In each test template, the true word and canary word have differing sentiment labels (+, 0, or \textminus), and only the true word may appear in a valid segmentation of the sentence. For each test template, a control template is formed by manually generating a locally unambiguous paraphrase of the test template's test site.

Paradigms are parameterized by the following properties of the test template.

\paragraph{Branching Structure.} Let $x_1x_2x_3$ denote the test site of a paradigm's test template. We say that this paradigm is \textit{left-branching} if $x_1x_2$ is the true word and $x_2x_3$ is the canary word. Otherwise, we say that this paradigm is \textit{right-branching.}

\paragraph{Sentiment.} The \textit{true sentiment} of a paradigm is the sentiment label of its true word, while the \textit{canary sentiment} is the sentiment label of its canary word. We represent the true sentiment $t$ and canary sentiment $c$ of a paradigm using the notation $t$/$c$.

\begin{figure}
    \centering
    \footnotesize
    \textbf{Left-Branching, +/\textminus} (Canary Word: 心机 `caution')
    
    \vspace{1em}
    \begin{tabular}{r c c c c}
        \textit{Test:} & person & \underline{留}\colorbox{nyuviolet!30}{\underline{心}} & \colorbox{nyuviolet!30}{\underline{机}}动的 & noun  \\
        & & takes & the & \\
        & & note of & motorized \\[1em]
        \textit{Control:} & person & \underline{留意} & \underline{机}动的 & noun\\
        & & takes & the & \\
        & & note of & motorized
    \end{tabular} 

    \vspace{1em}
    \textbf{Right-Branching, \textminus/+} (Canary Word: 留心 `to take note')
    
    \vspace{1em}
    \begin{tabular}{r c c c c c }
        \textit{Test:} & person & \colorbox{nyuviolet!30}{\underline{留}} &\colorbox{nyuviolet!30}{\underline{心}} \underline{机} & 处理 & noun  \\
        & & with & caution & treats & \\[1em]
       \textit{Control:} & person & \underline{留} & \underline{计谋} & 处理 & noun  \\
      &  & with & shrewdness & treats &  
    \end{tabular} 
    \vspace{.25em}
    
    \caption{Examples of paradigms in ERAS. Each paradigm consists of two templates, one for test sentences and one for control sentences. Each paradigm also has a branching structure (left or right) and a true and canary sentiment value (+/\textminus, +/0, \textminus/0, or \textminus/+). Templates contain \textit{slots} belonging to seven possible types: concept, entity, modifier, noun, object, person, or verb. The templates shown here contain one person slot and one noun slot.}
    \label{fig:paradigms}
\end{figure} 

\subsection{Explicit vs.\ Implicit Segmentation Errors}

ERAS measures susceptibility to two kinds of segmentation errors: \textit{explicit} and \textit{implicit}. An explicit segmentation error occurs when a model outputs word segmentation information that parses the canary word as a word. An implicit segmentation error occurs when a model does not output word segmentation information, but nonetheless behaves as though the canary word were parsed as a word. We detect implicit segmentation errors in \autoref{sec:sentiment}, where we evaluate sentiment analysis models on ERAS by testing the null hypothesis that sentiment predictions for ERAS sentences are not influenced by their paradigms' canary sentiment labels.

\subsection{Dataset Construction}
\label{sec:dataset-construction}

ERAS was constructed via the following steps.

\paragraph{Creation of Templates.} We form an initial list of test sentence test sites by extracting all word pairs of the form $(x_1x_2, x_2x_3)$ from the ANTUSD lexicon, where $x_1, x_2, x_3$ are single characters, such that $x_1x_2$ and $x_2x_3$ have different sentiment labels. For each word pair, we manually generate two Mandarin Chinese sentences subject to the following criteria: (1) both sentences contain the test site $x_1x_2x_3$, (2) both sentences are locally ambiguous at the test site and nowhere else, and (3) one sentence is left-branching and the other is right-branching. We discard all the word pairs for which we were unable to generate the two sentences, and we discard sentences for which the true sentiment label is 0.\footnote{As explained in \autoref{sec:human-eval}, we do not allow true sentiment labels to be 0 because such paradigms are not compatible with our human evaluation methodology.} The remaining 39 sentences are manually converted into test templates by replacing certain content words with one of the following \textit{slots}: concept, entity, modifier, noun, object, person, or verb. For each test template, we construct the corresponding control template by paraphrasing the test site without local ambiguity.

\paragraph{Sentiment Labeling.} ANTUSD provides sentiment scores on a continuous scale from \textminus 1 (negative) to 1 (positive), as well as categorical labels (positive, negative, or neutral) from each annotator. We consider a word to be labeled as + (resp.\ \textminus) if its sentiment score is at least .4 (resp.\ less than \textminus.4), and more annotators labeled the word as positive than negative (and \textit{vice versa}). We consider a word to be labeled as 0 if it meets the following criteria: its sentiment score is 0, the majority of annotators labeled it as \textit{neutral}, and at least two annotators marked it as a ``non-sentiment word.''

\paragraph{Slot Filling.} Each of the seven slot types (concept, entity, modifier, noun, object, person, verb), is associated with a manually constructed word list. To generate test and control sentences, we present each template to \texttt{bert-base-chinese}\footnote{\url{https://huggingface.co/google-bert/bert-base-chinese}} \citep{devlinBERTPretrainingDeep2019}, with all slots masked out. We then fill the slots iteratively from left to right, at each step using all words from the appropriate word list whose masked language modeling probability scores exceed a certain threshold.\footnote{The threshold is .74 for the first slot in each template, and for each subsequent slot, the threshold is increased by .05.}

\paragraph{Conversion to Traditional Characters.} ERAS is constructed using simplified characters. We create a traditional-character version of ERAS  using the \texttt{chinese-converter} Python package.\footnote{\url{https://pypi.org/project/chinese-converter/}}

\subsection{Human Evaluation}
\label{sec:human-eval}

Our setup for detecting implicit segmentation errors using ERAS is based on the assumption that both test and control sentences have the sentiment value given by the paradigm's true sentiment, but that test sentences are treated as having the canary sentiment if and only if they are segmented incorrectly due to garden path errors. In order to verify this, a sample of 20 sentence pairs from each paradigm was evaluated by 5 native speakers of Mandarin Chinese, with each annotator evaluating 4 sentence pairs per paradigm.  We assume that (1) the sentiment of control sentences is given by their paradigms' true sentiment labels and (2) annotators do not make garden path errors while reading test sentences,\footnote{In order to ensure this, we give annotators unlimited time to complete the task, and do not require them to label examples in order.} and we seek to verify that the sentiment of each test sentence matches that of its corresponding control sentence.

\paragraph{Annotation Task.} Annotators were asked to rank sentence triples in order of sentiment polarity. For each test--control pair $(t, c)$, annotators were presented with triples of the form $\lbrace t, c_1, c_2 \rbrace$ and $\lbrace c, c_1, c_2 \rbrace$, where $c_1$ and $c_2$ are control sentences. Each annotator ranked 208 sentence triples, the minimum necessary to ensure that all test--control pairs assigned to the annotator are represented.

\paragraph{Evaluation.} We assume that any sentence that is ranked more positively (resp.\ negatively) than a control sentence with a label of + (resp.\ \textminus) also has a label of + (resp.\ \textminus). Therefore, we say that a test--control pair $(t, c)$ with a true sentiment of + (resp.\ \textminus) is \textit{annotated correctly} if $t$ is ranked at least as positively (resp.\ negatively) as $c$, relative to $c_1$ and $c_2$. Because we have ensured in \autoref{sec:dataset-construction} that true sentiment labels can only be + or \textminus, we do not need to consider cases where the control sentence has a sentiment of 0; in such cases, unless $t$ has exactly the same ranking as $c$, we cannot determine whether or not $t$ has a sentiment of 0.

\paragraph{Result.}  We calculate a \textit{human accuracy} of 91.3, defined as the percentage of test--control pairs that have been annotated correctly. For 81.5\% of pairs, test and control sentences received the exact same sentiment ranking. This confirms that test and control sentences have the same sentiment value in the absence of garden path errors. We use this result to calculate a human baseline in \autoref{sec:sentiment}.

\section{Explicit Garden Path Errors}
\label{sec:word-segmentation}

\begin{table*}
    \centering
    \footnotesize
    \begin{tabular}{l c | c c c | c c c | c c c}
        \toprule 
        & & \multicolumn{3}{c|}{\textbf{Overall}} & \multicolumn{3}{c|}{\textbf{Left-Branching}} & \multicolumn{3}{c}{\textbf{Right-Branching}} \\
        & & Test & Control & Diff. & Test & Control & Diff. & Test & Control & Diff. \\\midrule
        \multicolumn{2}{l|}{\textit{Transformer Models}} & & & & & & & & & \\
        \hspace{1em}AS & (\texttt{hant}) & 82.3 & 97.4 & 15.1 & 91.4 & \textbf{100.0} & 8.6 & 76.6 & 89.6 & 13.0 \\
        \hspace{1em}CityU & (\texttt{yue}) & 81.8 & \textbf{99.3} & 17.5 & 83.2 & \textbf{100.0} & 16.8 & \textbf{92.0} & \textbf{99.8} & 7.8 \\
        \hspace{1em}PKU & & 84.4 & 94.9 & 10.5 & 94.2 & \textbf{100.0} & 5.8 & 82.4 & 84.6 & 2.2 \\
        \hspace{1em}MSR & & \textbf{88.6} & 94.4 & \textbf{5.8} & \textbf{95.8} & \textbf{100.0} & \textbf{4.2} & 88.3 & 89.6 & \textbf{1.3} \\\midrule 
        \multicolumn{2}{l|}{\textit{Non-Neural Models}} & & & & & & & & & \\
        \hspace{1em}Jieba & & 68.9 & \textbf{92.3} & 23.4 & 71.2 & 92.6 & 21.4 & 65.8 & \textbf{91.9} & 26.1 \\
        \hspace{1em}PKUSEG & & \textbf{75.4} & 90.0 & \textbf{14.6} & \textbf{78.4} & \textbf{95.5} & \textbf{17.1} & \textbf{71.2} & 82.1 & \textbf{10.9} \\\midrule
        \multicolumn{2}{l|}{\textit{Baseline}} & & & & & & & & & \\
        \hspace{1em}MaxMatch & & 64.1 & 100.0 & 35.9 & 82.6 & 100.0 & 17.4 & 37.5 & 100.0 & 62.5 \\\bottomrule
    \end{tabular}
    \caption{Overall test and control accuracies on ERAS for word segmentation models trained on traditional Mandarin (\texttt{hant}), traditional Cantonese (\texttt{yue}), or simplified Mandarin corpora (all others). Results are reported for the entire ERAS dataset (``Overall''), along with results for left-branching and right-branching paradigms only. The MSR model is least susceptible to garden path errors. All models except for CityU perform better on left-branching test sentences than right-branching test sentences.}
    \label{table:word-segmentation-results}
\end{table*}

Our first experiment evaluates word segmentation models according to their vulnerability to explicit segmentation errors due to garden path phenomena. Under the hypothesis that word segmentation models may segment canary words as words, we predict that test-sentence test sites will be segmented less accurately than control-sentence test sites.

\subsection{Experimental Setup}

We test word segmentation models by using them to segment all sentences in the ERAS dataset. We evaluate models based on whether test sites are segmented correctly; we do not measure how well models segment other parts of the sentence. We consider a left-branching sentence with test site $x_1x_2x_3$ to be \textit{segmented incorrectly} if a word boundary is placed between $x_1$ and $x_2$, but not between $x_2$ and $x_3$. A right-branching sentence is segmented incorrectly if a word boundary is placed between $x_2$ and $x_3$, but not between $x_1$ and $x_2$.

\paragraph{Evaluation.} For each paradigm, we define a model's \textit{paradigm test accuracy} to be the percentage of test sentences within that paradigm that the model segments correctly, and we define the model's \textit{overall test accuracy} to be the mean of the model's paradigm test accuracy across all paradigms. \textit{Paradigm control accuracy} and \textit{overall control accuracy} are defined analogously.

\paragraph{Models.} We fine-tune \texttt{bert-\allowbreak base-\allowbreak chinese} models on the following word segmentation benchmarks: AS, CityU, PKU, and MSR \citep{emersonSecondInternationalChinese2005}. Each of our models performs close to state of the art results reported by \citet{huangFastAccurateNeural2020} and \citet{tianImprovingChineseWord2020}.\footnote{Our AS, CityU, PKU, and MSR models obtain F1 scores of 96.7, 97.8, 96.5, and 98.3 on their respective test sets, compared to state of the art F1 scores of 96.6, 97.9, 96.6, and 98.4, respectively.} We evaluate the AS and CityU models on the traditional-character version of ERAS, and PKU and MSR on the simplified-character version. In addition to our fine-tuned models, we also evaluate the Jieba\footnote{\url{https://github.com/fxsjy/jieba}} and PKUSEG \citep{luoPKUSEGToolkitMultiDomain2022}\footnote{\url{https://github.com/lancopku/pkuseg-python}} non-neural word segmentation models.

\paragraph{Baseline.} Our baseline for this experiment is the MaxMatch algorithm \citep{changOptimizingChineseWord2008}, which iterates through a text from left to right and greedily segments the longest string starting at the current position that exactly matches at least one word from the MSR training corpus.

\subsection{Results}

As reported in the first column of \autoref{table:word-segmentation-results}, all of our models achieve a lower accuracy on test sentences than control sentences. This confirms our hypothesis that garden paths serve as a source of explicit word segmentation errors. We find that the MSR and PKU models are the least susceptible to garden path errors, in the sense that they exhibit the smallest gap in performance between test and control sentences. By that metric, the MaxMatch baseline and the Jieba model are the most susceptible to garden path errors. The non-neural PKUSEG model slightly outperforms the AS and CityU models in terms of test--control gap, but this is largely because PKUSEG achieves the lowest overall accuracy on control sentences.

\paragraph{Left-Branching Bias.} The other columns of \autoref{table:word-segmentation-results} report overall accuracies for left- and right-branching paradigms, separately. These results reveal a bias in our models for producing left-branching segmentations, in the sense that all of our models perform better on left-branching sentences than right-branching sentences. The only exception to this trend is that CityU performs better on right-branching test sentences than left-branching test-sentences. MaxMatch explicitly incorporates this bias by iterating through the text from left to right; but since the other models do not incorporate an explicit concept of directionality, we assume that the left-branching bias in these models is learned from the training corpus.

\section{Implicit Garden Path Errors}
\label{sec:sentiment}

Our second experiment evaluates sentiment analysis models according to their vulnerability to implicit segmentation errors due to garden path effects. Our hypothesis is that the sentiment polarity of canary words will contribute to the sentiment labels assigned by the models to test sentences, even though these words should, in theory, be excluded from the semantics of these sentences. We therefore predict that, for example, a sentiment analysis model should label a test sentence with a negative-sentiment canary word as being ``more negative'' on average than the corresponding control sentence. We also predict that providing word boundary information during pre-training or fine-tuning will improve a sentiment model's performance according to the aforementioned metric.

\subsection{Experimental Setup}

We test binary sentiment analysis models, which produce a single logit representing predicted sentiment polarity, by having them predict the sentiment of each sentence in ERAS. Our evaluation of these models is analogous to the paradigm described in \autoref{sec:human-eval}, which we used for our human evaluation of ERAS's sentiment properties. To that end, we consider a +/\textminus\ or +/0 sentence pair to be \textit{classified incorrectly} if its test sentence receives a lower sentiment score than its control sentence. A \textminus/+ or \textminus/0 sentence pair is classified incorrectly if its test sentence receives a higher sentiment score than its control sentence.

\paragraph{Evaluation.} We define the \textit{paradigm accuracy} of a sentiment analysis model on a paradigm to be the percentage of sentence pairs in that paradigm that are classified correctly by the model. The \textit{overall accuracy} of a model on ERAS is the mean of that model's paradigm accuracies across all paradigms.

\begin{figure}
    \centering
    \footnotesize
    \textbf{Occluded Test Sentence}
    
    \vspace{1em}
    \begin{tabular}{c c c c c }
        学生 & \colorbox{nyuviolet!30}{\underline{[MASK]}} &\colorbox{nyuviolet!30}{\underline{心}} \underline{机} & 处理 & 友人  \\
        the  & & caution & treats & his  \\
        student & & & & friends
    \end{tabular} 
    \vspace{.25em}
    
    `The student treats his friends with caution.'

    \vspace{1em}
    \textbf{Occluded Control Sentence} 
    
    \vspace{1em}
    \begin{tabular}{c c c c c }
        学生 & \underline{[MASK]} & \underline{计谋} & 处理 & 友人  \\
        the & & shrewdness & treats & his \\
        student & & & & friends
    \end{tabular} 
    \vspace{.25em}
    
    `The student treats his friends shrewdly.'
    
    \caption{Our occlusion study ablates canary words by masking out (``occluding'') the character in the test site not belonging to the true word.}
    \label{fig:occlusion}
\end{figure} 

\paragraph{Causal Analysis.} We expect that a model that is completely immune to implicit garden path errors would make no systematic distinctions in sentiment between test and control sentences. Such a model would achieve an overall accuracy of 50\%. When a model achieves an overall accuracy of 50\%, however, we do not know whether this is because the model did not make any garden path errors, or whether it is because the model made some number of garden path errors while also making errors in the opposite direction. For this reason, we perform an \textit{occlusion study} \citep{zeilerVisualizingUnderstandingConvolutional2014} in order to determine what percentage of ERAS misclassifications were caused by garden path errors.

As illustrated in \autoref{fig:occlusion}, our occlusion study involves masking out the 1st character in right-branching test sites and the 3rd character in left-branching test sites, effectively ablating the canary word in test sentences. We assume that an incorrect ERAS classification is caused by an implicit garden path error if ablating the canary word causes the model's control and test outputs to become more similar to one another. Following \citet{balkirNecessitySufficiencyExplaining2022}, we define \textit{necessity} to be the percentage of ERAS errors that are caused by implicit garden path errors according to our occlusion test, and \textit{sufficiency} to be the percentage of implicit garden path errors detected by our occlusion test that result in an ERAS misclassification. We calculate the \textit{garden path error rate} (GPER) as follows:
\[
\text{GPER} = (1 - \text{overall accuracy}) \cdot \text{necessity.}
\]
The GPER measures the percentage of ERAS examples for which the model makes an implicit garden path error, leading to a misclassification.

\begin{table*}[ht]
    \footnotesize
    \centering
    \begin{tabular}{l | c c | c c | r r r r | c c}
        \toprule 
        & \multicolumn{2}{c|}{\textbf{Overall}} & \multicolumn{2}{c|}{\textbf{Occlusion}} & \multicolumn{4}{c|}{\textbf{Control \textminus\ Test Sentiment}} & \multicolumn{2}{c}{\textbf{Test Set}} \\
        & GPER & Acc. & Necc. & Suff. & +/\textminus & +/0 & \textminus/0 & \textminus/+ & ASAP & ChnSentiCorp \\\midrule 
        \textit{Off-the-Shelf Models} & & & & & & & & \\
        \hspace{1em}RoBERTa-110M  & 45.4 & 41.7 & 77.9 & 89.4 & 6.3 & 4.9 & \textbf{2.3} & \textminus 18.7 & 97.8 & 96.6 \\
        \hspace{1em}RoBERTa-330M  & \textbf{38.0} & \textbf{52.2} & 79.6 & \textbf{50.3} & \textbf{1.4} & \textbf{.9} & 6.0 & \textbf{\textminus 2.2} & 97.9 & 96.7 \\
        \hspace{1em}Megatron-1.3B & 41.8 & 43.8 & \textbf{74.3} & 68.7 & 2.1 & 13.6 & 8.3 & \textminus 15.9 & \textbf{98.1} & \textbf{97.0} \\\midrule
        \textit{Fine-Tuned Models} & & & & & & & & \\
        \hspace{1em}Sentiment Only & 36.0 & \textbf{49.5} & 71.3 & 74.9 & \textbf{\textminus 3.3} & \textbf{\textminus 1.9} & \textbf{.3} & \textminus 27.6 & \textbf{97.6} & 95.4 \\
        \hspace{1em}WWM & 32.6 & 50.8 & 66.2 & 67.7 & 5.5 & \textminus 5.0 & \textminus 6.8 & \textminus 11.0 & 97.5 & 95.1 \\
        \hspace{1em}CWS & 23.7 & 41.3 & \textbf{40.3} & 45.9 & \textminus 9.7 & 9.2 & \textminus 10.9 & \textminus 20.3 & 97.2 & \textbf{95.5} \\
        \hspace{1em}WWM+CWS & \textbf{18.8} & 59.8 & 46.8 & \textbf{34.5} & \textminus 5.9 & \textminus 6.5 & \textbf{.3} & \textbf{.6} & 97.0 & \textbf{95.5}
        \\\midrule
        \textit{Baseline} & &   \\
        \hspace{1em}Humans & 8.7 & 50.6 & \\\bottomrule
    \end{tabular}
    \caption{Garden path error rates (GPER, smaller is better) and overall accuracies (closer to 50 is better) attained by our sentiment analysis models, compared to a human baseline. Test set accuracy (higher is better) and occlusion study results (lower is better) are also reported. ``Control \textminus\ Test Sentiment'' reports the average difference in probability scores for the + class between control and test sentences (closer to 0 is better), where positive values for +/\textminus\ and +/0 and negative values for \textminus/+ and \textminus/0 constitute evidence of implicit garden path errors. Among the off-the-shelf models, RoBERTa-330M is least susceptible to garden path errors. GPER results indicate that a combination of whole word masking and joint training on word segmentation (WWM+CWS) is effective in reducing susceptibility to garden path errors, though the accuracy and test set results suggest that the latter makes the model more susceptible to errors caused by other factors.}
    \label{table:sentiment-results}
\end{table*}

\paragraph{Models.} We evaluate three off-the-shelf sentiment analysis models from the Erlangshen family \citep{zhangFengshenbangBeingFoundation2023}: RoBERTa-110M\footnote{\url{https://huggingface.co/IDEA-CCNL/Erlangshen-Roberta-110M-Sentiment}}, RoBERTa-330M,\footnote{\url{https://huggingface.co/IDEA-CCNL/Erlangshen-Roberta-330M-Sentiment}} and Megatron-BERT-1.3B.\footnote{\url{https://huggingface.co/IDEA-CCNL/Erlangshen-MegatronBert-1.3B-Sentiment}} %

To test our prediction regarding word boundary supervision, we consider two methods for injecting word boundary information into a sentiment analysis model: \textit{whole word masking} \citep{cuiPreTrainingWholeWord2021a}, where only full words are masked during masked language model pre-training, and joint fine-tuning on word segmentation. We fine-tune sentiment analysis models with whole word masking (``WWM''), joint training on word segmentation (``CWS''), and a combination of these two techniques (``WWM+CWS''), and compare these models to a model fine-tuned on sentiment analysis only without whole word masking (``Sentiment Only''). Our WWM and WWM+CWS models are fine-tuned from the \texttt{chinese-\allowbreak bert-\allowbreak wwm-\allowbreak ext} model,\footnote{\url{https://huggingface.co/hfl/chinese-bert-wwm-ext}} which was pre-trained with whole word masking, while our CWS and Sentiment Only models are fine-tuned from \texttt{bert-\allowbreak base-\allowbreak chinese}. Our CWS and WWM+CWS models are jointly fine-tuned on the MSR dataset.

All of our models use character-level tokenization, and none of them incorporate an explicit word segmentation preprocessing step, apart from whole word masking or joint tuning on word segmentation.

\paragraph{Baseline.} We compare our models against the results of the human evaluation described in \autoref{sec:human-eval}. The human accuracy score of 91.3 reported there is not directly comparable to accuracies for sentiment analysis models, however, because the sentiment values generated by the annotators are discrete rather than continuous. In order to make the human evaluation results comparable to the results of this experiment, we assume that half of the 81.5\% of ERAS sentence pairs where test and control sentences received the same sentiment ranking would be annotated correctly in the continuous setting, and that all misclassifications are caused by garden path errors. This leads to a human baseline accuracy of 50.6 and GPER of 8.7.

\subsection{Results}

The results of our implicit garden path error experiment are shown in \autoref{table:sentiment-results}. The RoBERTa-330M model achieves the best GPER among the off-the-shelf models, and our WWM+CWS model achieves the best GPER among our fine-tuned models. The RoBERTa-330M model also achieves the best overall accuracy among the off-the-shelf models, but the WWM+CWS model achieves the worst overall accuracy across all models. Instead, the Sentiment Only model achieves the best overall accuracy, outperforming the human baseline.

\paragraph{Error Analysis.} The right-hand side of \autoref{table:sentiment-results} (``Control \textminus\ Test Sentiment'') shows the average difference in sentiment scores assigned by our models to control vs.\ test sentences, where scores range from 0 (most likely to be \textminus) to 100 (most likely to be +). We would ideally like these numbers to be as close to 0 as possible, since this would indicate that the model makes no systematic distinction between control and test sentences. This error analysis reveals that for all models except RoBERTa-330M and WWM+CWS, the \textminus/+ paradigms result in the greatest discrepancy in output between control and test sentences. Indeed, the three models with ERAS accuracies above 50 are also those with the smallest control--test differences for \textminus/+ paradigms. This suggests that \textminus/+ sentences are the ones for which our models are most likely to commit implicit garden path errors.

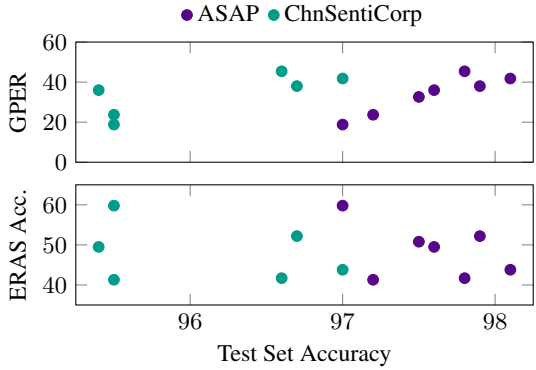
\begin{figure}
    \centering
    \footnotesize
    \begin{tikzpicture}
        \begin{axis}[%
            width=.475\textwidth, height=1.25in,
            xmin=95.25, xmax=98.25,
            ymin=0, ymax=60,
            xticklabel=\empty,
            ylabel=GPER,
            legend style={at={(0.5,1.05)},anchor=south,legend columns=2,draw=none},
            scatter/classes={
                a={mark=*,draw=nyuviolet,fill=nyuviolet},
                b={mark=*,draw=nyuteal,fill=nyuteal}
            }]
            \addplot[scatter,only marks,scatter src=explicit symbolic]%
                table[meta=label] {
                    x y label
                    97.8 45.4 a
                    97.9 38.0 a
                    98.1 41.8 a
                    97.6 36.0 a
                    97.5 32.6 a
                    97.2 23.7 a
                    97.0 18.8 a
                };
            \addlegendentry{ASAP\,}
            \addplot[scatter,only marks,scatter src=explicit symbolic]%
                table[meta=label] {
                    x y label
                    96.6 45.4 b
                    96.7 38.0 b
                    97.0 41.8 b
                    95.4 36.0 b
                    95.1 32.6 b
                    95.5 23.7 b
                    95.5 18.8 b
                };
            \addlegendentry{ChnSentiCorp}
        \end{axis}
    \end{tikzpicture} \\
    \begin{tikzpicture}
        \begin{axis}[%
            width=.475\textwidth, height=1.25in,
            xmin=95.25, xmax=98.25,
            ymin=35, ymax=65,
            ylabel={ERAS Acc.},
            xlabel={Test Set Accuracy},
            scatter/classes={
                a={mark=*,draw=nyuviolet,fill=nyuviolet},
                b={mark=*,draw=nyuteal,fill=nyuteal}
            }]
            \addplot[scatter,only marks,scatter src=explicit symbolic]%
                table[meta=label] {
                    x y label
                    97.8 41.7 a
                    97.9 52.2 a
                    98.1 43.8 a
                    97.6 49.5 a
                    97.5 50.8 a
                    97.2 41.3 a
                    97.0 59.8 a
                };
            \addplot[scatter,only marks,scatter src=explicit symbolic]%
                table[meta=label] {
                    x y label
                    96.6 41.7 b
                    96.7 52.2 b
                    97.0 43.8 b
                    95.4 49.5 b
                    95.1 50.8 b
                    95.5 41.3 b
                    95.5 59.8 b
                };
        \end{axis}
    \end{tikzpicture}
    \caption{Test set accuracy is correlated with GPER, but exhibits no obvious relationship with ERAS accuracy.}
    \label{fig:asap-occlusion}
\end{figure}

\paragraph{Word Boundary Supervision.} The GPER results show that both whole word masking and joint training on word segmentation are effective methods for mitigating susceptibility to implicit garden path errors, and that using both techniques in combination results in the best GPER. However, the CWS and WWM+CWS models also have the poorest performance in terms of ERAS overall accuracy as well as ASAP test set accuracy. This suggests that joint training on word segmentation is effective in reducing susceptibility to garden path errors, but makes the model more susceptible to other types of errors, with the costs of the latter outweighing the benefits of the former. On the other hand, the use of whole word masking does not exhibit this tradeoff, though it only enables a modest reduction in GPER.

\paragraph{Test Accuracy and GPER.} \autoref{fig:asap-occlusion} reveals a correlation between test set accuracy and GPER: better-performing sentiment analysis models are more susceptible to implicit garden path errors. No such relationship is observed between test set accuracy and ERAS overall accuracy, however. Thus, the increased susceptibility to garden path errors for better-performing models is offset by other factors that influence ERAS overall accuracy.

\section{Discussion and Related Work}

\paragraph{Left-Branching Bias in Humans.} Much work has been done in psycholinguistics on the human processing of Chinese morphological garden paths \citep{inhoffEyeMovementsIdentification2005,huangEarlyNotOverwhelming2020,huangPriorContextInfluences2021,tongCompetitionOverlappingAmbiguous2023,huangEffectsLexicalSentencelevel2024}. \citet{huangEarlyNotOverwhelming2020}, in particular, identify a human left-branching bias, similar to the left-branching bias observed for word segmentation models in \autoref{sec:word-segmentation}. They find that human readers spend less time looking at left-branching morphological garden paths than right-branching ones, all else equal, indicating that the former are easier to process than the latter. On the other hand, \citet{liaoLargeLanguageModels2024} did not observe a left-branching bias when instructing GPT-3.5 models to perform word segmentation.

\paragraph{Garden Paths in Word Segmentation.} Rule-based, statistical, and non-neural machine learning methods have been proposed for correctly segmenting morphological garden paths in Chinese word segmentation \citep{sunAmbiguityResolutionChinese1995,hanResolving2001,liUnsupervisedTrainingOverlapping2003,xiongNewMachineLearning2007,gaoDealingChineseOverlapping2009}. These techniques mostly involve first identifying locally ambiguous substrings, possibly with the help of a lexicon of locally ambiguous substrings, and then choosing a segmentation for each of those substrings based on the surrounding context. Corpus analyses have found that morphological garden paths comprise roughly 4\% of Chinese text \citep{qiaoStatisticalPropertiesOverlapping2008,yenUsageStatisticalCues2012}.

\paragraph{Targeted Evaluation in NLP.} Minimal pair benchmarks have been used to evaluate language models' compliance with morphosyntactic constraints such as subject--verb agreement or negative polarity item licensing \citep{linzenAssessingAbilityLSTMs2016,marvinTargetedSyntacticEvaluation2018,warstadtBLiMPBenchmarkLinguistic2020}. \citet{fuRethinkCWSChineseWord2020} propose an error analysis framework that compares Chinese word segmentation models' performance across inputs with differing values for various features, such as sentence length or word frequency.

\section{Conclusion}

This paper has proposed a benchmark, ERAS, that detects garden path errors in Chinese NLP models, with or without an explicit word segmentation step in the pipeline. Using ERAS, we have observed that both word segmentation models and sentiment analysis models are vulnerable to garden path errors. In both cases, models seem to inherit this vulnerability from their training distributions. The MSR segmentation corpus most effectively defends against garden path errors, while the ASAP dataset seems to exacerbate susceptibility to garden path errors. We also find that garden path errors are driven by particular kinds of inputs: left-branching inputs in the case of word segmentation, and \textminus/+ inputs in the case of sentiment analysis.

It has been argued that character-level tokenization is superior to word-level tokenization for Chinese neural NLP models \citep{liWordSegmentationNecessary2019}, and indeed, most Transformer language models for Chinese use character- or byte-level tokenization. Our findings from \autoref{sec:sentiment}, however, have shown that a total lack of word boundary information makes a model highly vulnerable to implicit garden path errors. We have shown that injecting word boundary information into the model through whole word masking or joint training on word segmentation can significantly reduce implicit garden path errors, though care must be taken to avoid degradations in model performance. We leave the development of such measures to future work.

\section{Limitations}

ERAS can only be used for two kinds of models: word segmentation models and sentiment analysis models. In particular, our mode of evaluation for sentiment analysis models requires access to the model's logits, and therefore it is not compatible with large language models whose logits are not available to the user. 

ERAS is also a large, synthetically generated dataset, consisting of minimal pair paradigms. Large synthetic minimal pair datasets are known to have the following limitations. First, large datasets are intrinsically impossible to evaluate fully \citep{benderDangersStochasticParrots2021}, and synthetic minimal pair datasets are often found to contain examples that do not conform to the intended properties of the datasets \citep{blodgettStereotypingNorwegianSalmon2021}. Second, minimal pair datasets have been criticized for lacking structural diversity \citep{javiervazquezmartinezEvaluatingNeuralLanguage2023}. ERAS is not an exception to either of these limitations.

\section{Ethical Considerations}

We are not aware of any risks or ethical concerns arising from the work described in this paper.

\bibliography{custom}

\appendix

\end{CJK}
\end{document}